\documentclass[runningheads]{llncs}
\usepackage{booktabs}
\usepackage{tabularx}
\usepackage{makecell}
\usepackage{amsmath}
\usepackage{placeins} 
\usepackage{hyperref} 
\usepackage[T1]{fontenc}
\usepackage{graphicx}
\usepackage{subcaption}
\usepackage{booktabs}
\usepackage{makecell}
\usepackage{caption}
\usepackage{xurl}      
\usepackage{hyperref}

\usepackage{array}

\usepackage{tcolorbox}
\begin{document}
\title{REFINE: Real-world Exploration of Interactive Feedback and Student Behaviour}
%
%
\author{Fares Fawzi\orcidID{0000-0002-5331-579X} \and
Seyed Parsa Neshaei\orcidID{0000-0002-4794-395X} \and
Marta Knezevic\orcidID{0009-0000-8588-0753} \and
Tanya Nazaretsky \orcidID{0000-0003-1343-0627} \and
Tanja Käser \orcidID{0000-0003-0672-0415}}


\authorrunning{Fawzi et al.}

\institute{%
EPFL, Lausanne, Switzerland\\
\email{\{firstname.lastname\}@epfl.ch}
}

%
%
\maketitle              
\begin{abstract}
Formative feedback is central to effective learning, yet providing timely, individualised feedback at scale remains a persistent challenge. While recent work has explored the use of large language models (LLMs) to automate feedback, most existing systems still conceptualise feedback as a static, one-way artifact, offering limited support for interpretation, clarification, or follow-up. In this work, we introduce REFINE, a locally deployable, multi-agent feedback system built on small, open-source LLMs that treats feedback as an interactive process. REFINE combines a pedagogically-grounded feedback generation agent with an LLM-as-a-judge-guided regeneration loop using a human-aligned judge, and a self-reflective  tool-calling interactive agent that supports student follow-up questions with context-aware, actionable responses. We evaluate REFINE through controlled experiments and an authentic classroom deployment in an undergraduate computer science course. Automatic evaluations show that judge-guided regeneration significantly improves feedback quality, and that the interactive agent produces efficient, high-quality responses comparable to a state-of-the-art closed-source model. Analysis of real student interactions further reveals distinct engagement patterns and indicates that system-generated feedback systematically steers subsequent student inquiry. Our findings demonstrate the feasibility and effectiveness of multi-agent, tool-augmented feedback systems for scalable, interactive feedback.
\end{abstract}

\section{Introduction}

Feedback is widely recognised as central to student learning, as it helps learners understand how their current performance aligns with expectations and what actions can improve it~\cite{PANADERO2022100416}. Effective formative feedback clarifies learning goals, supports learners in evaluating their work relative to those goals, and provides guidance for improvement \cite{hattie}. However, providing high-quality, personalised formative feedback at scale, particularly in large classes with limited instructor time, remains challenging.

Consequently, a growing body of work has explored using large language models (LLMs) to automate formative feedback generation to provide timely and personalised guidance at scale~\cite{Cao2025FirstDraft,Scarlatos2024RLFeedback,rashkin-etal-2025-help,10.1007/978-3-031-64299-9_6}. Although LLM-generated feedback is often fluent, specific, and comparable to human-authored feedback in controlled settings, prior studies show that it frequently fails to identify or prioritise the most important issues in students’ work \cite{rashkin-etal-2025-help,Cao2025FirstDraft,ChatGPTs-Decimal-Skills}. This suggests that surface-level fluency does not reliably translate into pedagogically effective feedback. 

To address these limitations, prior work has explored structured single-pass prompting grounded in theory, showing that LLMs can generate pedagogically plausible feedback but still lag expert teachers in prioritisation and actionable guidance \cite{STEISS2024101894,rashkin-etal-2025-help}. Subsequent work has adopted increasingly complex prompting-based designs, including multi-stage pipelines, in which feedback is generated, evaluated, and iteratively regenerated against pedagogical criteria \cite{Cao2025FirstDraft}. This line of work also explores techniques such as retrieval-augmented generation (RAG) \cite{rag} with chain-of-thought (CoT) \cite{cot} to support more accurate and factual feedback \cite{Cao2025FirstDraft}, or employs specialised agents and student performance signals to adapt feedback strategies~\cite{Partnering-with-AI}. However, these prompting-based approaches predominantly rely on large, closed-source LLMs accessed via commercial APIs, raising concerns around cost, privacy, reproducibility, and classroom deployability \cite{Cao2025FirstDraft,Partnering-with-AI}.

Complementary work has therefore explored optimisation-based methods to improve feedback validity using smaller, open-source models. These include preference-based approaches such as Direct Preference Optimisation (DPO) \cite{rafailov2023direct}, which leverage pairwise preferences from expert annotators \cite{2025.EDM.short-papers.166} or larger LLMs \cite{Scarlatos2024RLFeedback,nair-etal-2024-closing}. While effective, such pipelines typically require substantial preference data, repeated evaluation, and significant computational resources, limiting their practicality and scalability for classroom deployment\cite{Scarlatos2024RLFeedback,nair-etal-2024-closing}.

Most LLM-generated feedback is delivered as a static, one-way artefact after task completion \cite{Cao2025FirstDraft,Scarlatos2024RLFeedback,Optimizing-Educational-Feedback,2024.EDM-posters.99}, mirroring long-standing feedback practices that limit student engagement. However, evidence suggests that feedback quality alone is insufficient for learning gains unless learners actively engage with and use it \cite{10.1007/978-3-032-03870-8_33,Dawson02012019}. In static formats, students often struggle to interpret which aspects of their work are referenced, how comments relate to assessment criteria, or how suggested changes could be applied, frequently requiring dialogue and clarification to make sense of the feedback \cite{nicol2006principles,Bailey01042010,Yang01042013,carless2018feedbackliteracy,uzun2026studentsaskgenerativeai}. As a result, many LLM-based approaches risk reinforcing the gap between feedback generation and uptake, offering limited support for clarification, sense-making, and follow-up engagement.

In this paper, we introduce REFINE, a locally deployable multi-agent interactive feedback system built on small, open-source thinking language models. REFINE supports an interactive feedback process in which a student submits a solution, receives structured formative feedback grounded in established feedback theory, and then engages in follow-up dialogue to clarify and extend that feedback, enabling us to study how students interpret and act on feedback through their interactions. The system couples (1) a pedagogically grounded feedback generation agent that produces a structured feedback report and improves it through a human-aligned LLM-as-a-judge–guided iterative refinement loop, where feedback components are evaluated against a rubric derived from prior literature and selectively regenerated when criteria are not met, and (2) a tool-calling interactive feedback agent that answers student follow-up questions through closed-loop reasoning and tool use grounded in the generated feedback and course materials. 

Using REFINE, we investigate (RQ1) whether a multi-agent system built on small, open-source models can generate pedagogically aligned feedback and high-quality interactive responses, (RQ2) how students engage with feedback when follow-up questions are supported, and (RQ3) whether generated feedback content steers subsequent student dialogue. We evaluate REFINE in an undergraduate discrete mathematics course (in-class: 39 reports and conversations; exam preparation: 362 reports, 99 conversations). Results show that judge-guided refinement improves key pedagogical aspects of feedback while maintaining robust correctness in deployment. REFINE fine-tuning narrows the gap between an 8B model and GPT-5 on relevance and actionability, and student follow-up questions cluster into distinct engagement themes shaped by what the feedback highlights.

We provide our full implementation, and an online appendix on our GitHub repository: \url{https://github.com/epfl-ml4ed/REFINE}

\begin{figure}[t]
  \centering
  \includegraphics[width=\linewidth]{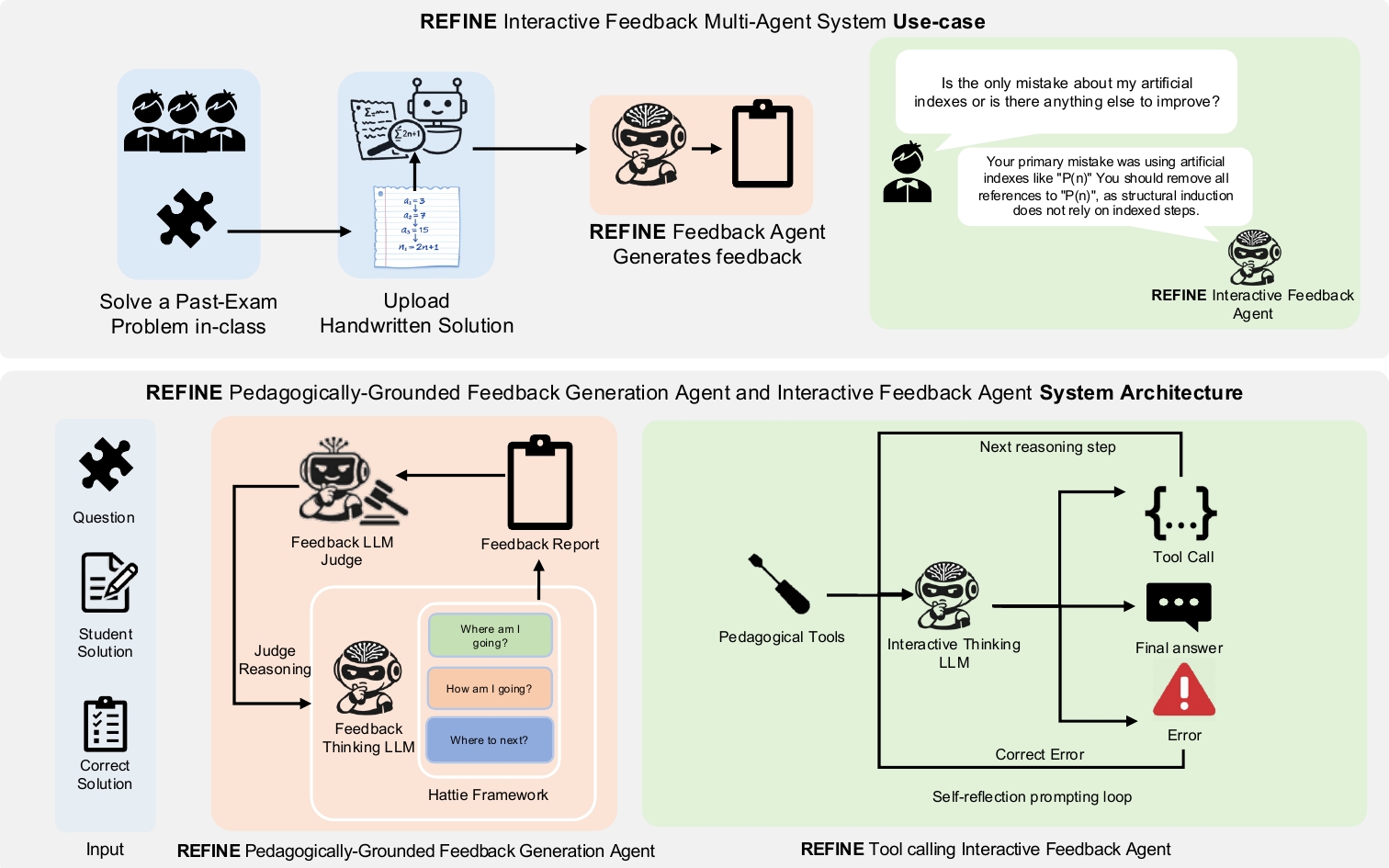}
\caption{\textbf{Interactive feedback workflow and system overview.}
\textbf{(Top)} Classroom workflow where students submit handwritten solutions, receive structured feedback, and ask questions. \textbf{(Bottom)} REFINE multi-agent system, where feedback is iteratively refined by a human-aligned LLM judge and paired with a trained tool-calling agent that answers via closed-loop, self-reflective reasoning.}
  \vspace{-4mm}
  \label{fig:methods_pipeline}
\end{figure}

\vspace{-2mm}
\section{REFINE: Multi-Agent Interactive Feedback System}
\vspace{-1mm}

Our goal is to enable interactive feedback with small open-source models by generating pedagogically grounded feedback and supporting follow-up through a tool-calling interactive feedback agent. REFINE supports this interactive workflow in which learners solve a task, upload a handwritten solution, receive structured formative feedback, and ask follow-up questions (Fig. \ref{fig:methods_pipeline}, top).
Uploaded handwritten work is transcribed using the Qwen2.5-VL-72B-Instruct vision\allowbreak--\allowbreak language model \cite{qwen25VL}, which students can review and edit  before submission. This workflow is implemented by a two-stage, multi-agent architecture (Fig. \ref{fig:methods_pipeline}, bottom) that separates feedback generation from interaction. In the first stage, a pedagogically grounded Feedback Generation Agent produces a structured feedback report aligned with established feedback theory \cite{hattie}, which is then refined through a generate–evaluate–revise loop guided by a human-aligned Feedback LLM Judge that assesses component-level pedagogical criteria and provides targeted revision rationales. In the second stage, the refined feedback report is passed to an Interactive Feedback Agent, which conditions on the student’s follow-up question and the feedback report to perform multi-step self-reflective reasoning, invoke pedagogical tools, and generate a response through multi-hop reasoning.

\vspace{-2mm}
\subsection{Pedagogically-Grounded Feedback Generation Agent}
\vspace{-2mm}
The pedagogically grounded Feedback Generation Agent comprises a Feedback Thinking LLM and a Feedback LLM Judge–guided refinement mechanism, which together generate and iteratively improve structured, theory-grounded feedback.

\vspace{1pt} \noindent \textbf{Feedback Structure and Pedagogical Objectives.}
The Feedback Thinking LLM produces structured, pedagogically grounded feedback intended to help students understand their performance and identify actions for improvement. Its design is grounded in established feedback theory \cite{hattie}, which organises formative feedback around the questions ``Where am I going?'', ``How am I going?'', and ``Where to next?'' Given a question $q$, a student solution $s$, and a correct solution $r$, the Feedback Thinking LLM generates a structured feedback report $f$, composed of $f_{\mathrm{cs}}$, $f_{\mathrm{task}}$, $f_{\mathrm{strat}}$, $f_{\mathrm{sr}}$, and $f_{\mathrm{p}}$. The component $f_{\mathrm{cs}}$ addresses ``Where am I going?'' by clarifying the task objectives and ``How am I going?'' by diagnosing the student’s current state relative to the task requirements. $f_{\mathrm{task}}$, $f_{\mathrm{strat}}$, and $f_{\mathrm{sr}}$ address ``Where to next?'' through task-level next steps, strategy-level guidance, and self-regulated learning support. The praise component $f_{\mathrm{p}}$ acknowledges effective aspects of the student’s work. This decomposition enforces coverage across multiple pedagogical levels while maintaining clear separation of feedback functions. All LLM prompts are provided in the GitHub repository.

\vspace{1pt} \noindent \textbf{Feedback LLM Judge-Guided Refinement}
\label{sec:iterative feedback generation}
Inspired by prior work on multi-stage feedback generation \cite{Cao2025FirstDraft}, feedback refinement follows an iterative generate, evaluate, and revise process guided by the Feedback LLM Judge.
The judge is a human-aligned LLM that evaluates feedback quality using a rubric adapted from prior work \cite{NAZARETSKY2026100533}, covering clarity, diagnosis of the student’s current state, task-level, strategy-level, and self-regulated next steps, and praise. For current-state and task-level next steps, we distinguish coverage (whether the component is addressed) from correctness (whether the analysis or recommendation is accurate and specific to student’s solution and task). This is implemented by embedding the rubric into a structured prompt that guides the model to reason jointly about the question, the student’s solution, and the generated feedback, and to produce a rubric-based judgment with an accompanying explanation. We validate the alignment of the LLM judge with this prompting by measuring agreement with expert annotations, reported as Cohen’s $\kappa$ in Section \ref{sec: offline exp}. The full judge prompts are provided in the GitHub repository. 
For a feedback instance $f$, the Feedback LLM Judge evaluates each component $f_k$ and returns a binary judgment $y_k \in \{0,1\}$; when $y_k=0$, it also provides a structured and targeted explanation $e_k$ describing the deficiency \cite{saha2025learning}. Only these targeted explanations are passed to the Feedback Thinking LLM, which revises the failing components while preserving those that already satisfy the rubric. This process repeats until all rubric components are satisfied or a predefined iteration limit is reached.

\subsection{Tool-Calling Interactive Feedback Agent}
\label{sec:interactive_feedback_agent_sys}
The Tool-calling Interactive Feedback Agent answers student follow-up questions by conditioning on the structured feedback report $f$ and invoking external pedagogical tools. An Interactive Thinking LLM performs multi-step, self-reflective reasoning to select tool calls, incorporate tool observations, recover from execution errors, and produce a final response that is evaluated by an Interactive Feedback Judge (Fig.~\ref{fig:methods_pipeline}, bottom). Given a student query $u$ and a structured feedback report $f$, the agent follows a closed-loop inference process with multi-step reasoning and tool use \cite{fawzi-etal-2025-scribe}. Let $\mathcal{T}$ denote a finite set of tools, and $o_i$ the observation returned by executing a tool call at step $i$. At each step $i = 0, \ldots, n$, the agent generates a reasoning state $r_i$ conditioned on the accumulated context $(u, f, r_{<i}, o_{<i})$ and proposes a tool call $t_i \in \mathcal{T}$, yielding either an observation $o_i$ or an execution error. If an error occurs, the agent revises the tool call. Otherwise, the resulting observation is incorporated into the context for the next step. Tool use continues until sufficient evidence is accumulated to produce a response $a$ or a predefined limit is reached, yielding a trajectory $(u, f, r_0, t_0, o_0, \ldots, r_n, a)$. Conditioning each step on the full interaction history enables adaptive multi-step reasoning, tool selection, and error recovery within a single inference loop. 
To support this behaviour, the interactive agent is fine-tuned on supervised interaction traces comprising student questions, intermediate reasoning steps, tool calls, tool outputs, and final responses. Training uses a two-stage LoRA procedure: the first stage trains initial reasoning and tool selection, and the second stage trains multi-step tool use and response generation, yielding a REFINE interactive feedback model. We refer to any base model fine-tuned with this procedure as a \(\langle\)base model\(\rangle\) REFINE model.

The tool set follows \cite{fawzi-etal-2025-scribe} and supports retrieval-based access to instructional content, prerequisite information, and behavioural reasoning to ground interactive feedback. Course resources (textbook, syllabus, lecture slides, and exercises) are converted to structured markdown using the Qwen2.5-VL-72B-Instruct model \cite{qwen25VL} and indexed with a hybrid retrieval scheme combining BM25 lexical retrieval with semantic search based on Qwen3-Embedding-0.6B embeddings \cite{qwen3technicalreport}. Prerequisite information is encoded via topic dependency maps inferred by prompting GPT-5 \cite{singh2025openaigpt5card} and validated by the course teaching team. Behavioural tools support explanation and hypothetical reasoning by comparing student queries against a fixed set of behavioural dimensions, namely Effort, Consistency, Proactivity, Assessment, and Regularity \cite{behave_dimensions}, and returning grounded explanations of how alternative behaviours may influence learning outcomes. The tools are implemented as functions that the LLM invokes via structured JSON calls. Full implementation details and schemas are provided in the GitHub repository

\section{Experimental Design}
We evaluated REFINE both offline on existing labelled datasets as well as through deployment with real students in both a controlled classroom study and an unconstrained exam-preparation setting. All three settings focused on a course in discrete mathematics at the undergraduate level.

\subsection{Offline Experiments}
\label{sec: offline exp}
We evaluated and trained the feedback generation agent and the interactive feedback agent on curated datasets before deployment.

\vspace{1pt} \noindent \textbf{Feedback Generation.} We evaluated feedback quality on $\mathcal{D}_{fb}$, an existing dataset of student solutions to an open-ended discrete mathematics proof task \cite{NAZARETSKY2026100533}. The dataset contains $177$ student solutions paired with reference solutions.

We compared the performance of three language models on $\mathcal{D}_{fb}$: Qwen3-8B REFINE, Qwen3-30B-A3B-Thinking-2507 (Qwen3-30B-Thinking), and GPT-5 as a gold-standard. Each model generated structured feedback for a given student solution and iteratively refined it via judge-guided regeneration until all pedagogical criteria were satisfied or a maximum of $20$ iterations was reached. To validate the reliability of the feedback judge and its underlying rubric, two expert annotators independently annotated a randomly sampled subset of $25$ generated feedback instances. Human–human agreement was high across all dimensions (Cohen’s $\kappa \geq 0.92$). We used a Qwen3-30B-Thinking model as the judge and evaluated it by running it five times per instance and comparing its judgments against expert annotations, yielding consistently high human–judge agreement (mean $\kappa = 0.95$), with per-dimension agreement ranging from $0.77 \pm 0.15$ (task-level next-step correctness) to $1.00 \pm 0.00$ (clarity, coverage, and praise).

\vspace{1pt} \noindent \textbf{Interactive Feedback Agent.}
We trained and evaluated the interactive feedback agent using the dataset from \cite{fawzi-etal-2025-scribe}, denoted $\mathcal{D}_{int}^{train}$ and $\mathcal{D}_{int}^{test}$. This dataset provides supervised interaction traces comprising approximately 7{,}000 student questions along with intermediate reasoning steps, tool calls, tool outputs, and final responses, and a held-out test set of $192$ student questions.

We finetuned a Qwen3-8B Thinking model on the $\mathcal{D}_{int}^{train}$ using two-stage LoRA finetuning (see Section~\ref{sec:interactive_feedback_agent_sys}), yielding Qwen3-8B REFINE. In preliminary analysis, we observed that the Qwen3-8B thinking-enabled base model outperformed instruction-tuned models of comparable size reported in prior SCRIBE experiments \cite{fawzi-etal-2025-scribe}. This motivated us to investigate whether further training on supervised interaction traces could refine and selectively surface the most informative internal reasoning behaviours of such models. We therefore constrain surfaced reasoning to a compact summary, motivated by findings that only a limited subset of reasoning steps meaningfully influence correctness, whereas exposing full reasoning traces adds little additional signal \cite{qian2025demystifying}.

To assess training, we compared performance on $\mathcal{D}_{int}^{test}$ against Qwen3-8B with reasoning disabled, Qwen3-8B Thinking, and GPT-5, isolating the contribution of internal reasoning generation. Generated responses were evaluated using a human-aligned Interactive Feedback Judge implemented with GPT-4.1 and following the evaluation rubric introduced in \cite{fawzi-etal-2025-scribe}, which measures Relevance (alignment with the student question), Actionability (provision of concrete guidance), Tool Relevance (appropriateness of selected tools), and Correctness (factual consistency with tool outputs and feedback). In addition to response quality, we analysed tool-use efficiency using interaction traces, measuring the number of reasoning-tool steps and tool calls required to reach a final response.


\subsection{Controlled Classroom Study}
\label{sec:study_method}
To evaluate REFINE in authentic use, we conducted a controlled classroom study during a regular session of a first-year undergraduate discrete mathematics course at EPFL, where 39 students (13 identified as female) used REFINE to solve a past exam question on structural induction. After uploading their solution and receiving the feedback report, students were required to ask at least three follow-up questions and then completed a post-interaction perception survey adapted from prior work \cite{NAZARETSKY2026100533,AI-or-Human}. The survey used a 5-point Likert scale to assess perceived correctness, relevance, clarity, and actionability of the feedback. All participants provided informed consent and the study was approved by the university ethics review board (Approval No. HREC000714 / 19.12.2025). Student interactions yielded two datasets, $\mathcal{D}_{study}^{feedback}$ and $\mathcal{D}_{study}^{int}$. Based on offline analyses, we used Qwen3-30B-Thinking for feedback generation and Qwen3-8B REFINE as the interactive agent. To ensure low latency in the live classroom deployment, where students interacted with the system in real time, regeneration was limited to a single iteration targeting current-state and task-next-step correctness. In this configuration, the selected models support interactive usage, achieving approximately ~20 tokens/s for Qwen3-8B REFINE on a single NVIDIA A100 GPU and ~21.5 tokens/s for Qwen3-30B-Thinking on two A100 GPUs.

\vspace{1pt} \noindent \textbf{Model Performance.} We performed a post-hoc validation of feedback generation and interactive feedback conversations on $\mathcal{D}_{study}^{feedback}$ and $\mathcal{D}_{study}^{int}$ using the human-aligned Feedback LLM Judge as well as the Interactive Feedback Judge. We verified human alignment of the judges again on a subset of the data, obtaining high inter-rater agreement ($\kappa \geq 0.83$) for both judges and all dimensions.

\vspace{1pt} \noindent \textbf{Student Engagement.} To analyse student engagement, we conducted a thematic analysis of student follow-up questions using interaction logs from $\mathcal{D}_{study}^{int}$. Two researchers first independently created themes from the collected data and subsequently discussed and refined them, identifying $15$ recurring themes grouped into five higher-level engagement categories. Table~\ref{tab:question-categories} summarises the categories, underlying themes, and their interpretation. The \textit{Task} category captures questions clarifying the task requirements, indicating difficulty interpreting what was being asked rather than engaging with solution content. The \textit{Solution} category focuses on correctness and completion, including attempts to repair parts of a submitted solution, requests for reassurance, or requests for a full model answer. \textit{Feedback}-focused questions reflect direct engagement with the feedback, such as asking for clarification, negotiating its interpretation, or asking for reassessment of the solution. The \textit{Additional Feedback} category captures questions going beyond the immediate task, such as advice on how to further improve or transfer learning to future problems. Finally, \textit{Off-topic} questions concern system usability or personal queries rather than the mathematical content. Both researchers then independently annotated all student questions, achieving substantial inter-rater agreement across questions and themes (Cohen’s $\kappa = 0.73$).

\begin{table}[t]
    \centering
    \setlength{\tabcolsep}{6pt}
        \caption{Question categories used to characterise student interactions with interactive feedback agent.}
    \resizebox{\textwidth}{!}{
    \begin{tabular}{p{0.18\linewidth} p{0.38\linewidth} p{0.38\linewidth}}
        \toprule
        \textbf{Category} & \textbf{Themes} & \textbf{Interpretation} \\
        \midrule
        \textit{Task} & Understanding task & Difficulty interpreting requirements \\
        \textit{Solution} & Repairing; reassurance; requesting a model answer & Focus on correctness and completion \\
        \textit{Feedback} & Interpreting; negotiating; elaboration; reassessment & Active engagement with feedback \\
        \textit{Additional Feedback} & Generalisation; seeking further improvements & Deepening and transfer \\
        \textit{Off-topic} & Usage questions; personal support; greeting/thanking & Usability / friction \\
        \bottomrule
    \end{tabular}
    }
    \label{tab:question-categories}
    \vspace{-2mm}
\end{table}

\vspace{1pt} \noindent \textbf{Feedback Uptake.} To investigate students' feedback uptake, we annotated all generated feedback reports and all student questions using the task grading rubric used by the teaching team in the past exam, consisting of basis verification (Basis), inductive hypothesis formulation (IH), and inductive step (Step). Inter-rater agreement between two expert annotators was moderate for feedback annotations (Cohen’s $\kappa = 0.665$) and for chat interaction annotations (Cohen’s $\kappa = 0.625$). In addition, a member of the teaching team graded each student solution using the same rubric, enabling analysis of alignment between feedback content, student questions, and task performance.

\vspace{-2mm}
\subsection{Unconstrained Exam-Preparation Use}
Finally, we deployed REFINE for students of the discrete mathematics course during final-exam preparation. In this unconstrained setting, students could freely access REFINE and choose any question from past exams over the last three years, spanning topics including logic, proofs, and algorithms, complexity, counting, and probability. We recorded all interactions, yielding an additional dataset $\mathcal{D}_{prep}^{feedback}$ containing $362$ feedback reports and $\mathcal{D}_{prep}^{int}$ containing $99$ student follow-up conversations. We used the same model configuration, refinement settings, and validation procedure as in the controlled classroom study (Section~\ref{sec:study_method}), using Qwen3-30B for feedback generation and Qwen3-8B REFINE as the interactive feedback agent. Post-hoc validation on a sampled subset of $\mathcal{D}_{prep}^{feedback}$ and $\mathcal{D}_{prep}^{int}$ showed high inter-rater agreement for both the Feedback Judge and the Interactive Feedback Judge across all dimensions ($\kappa \geq 0.75$).

\vspace{-2mm}
\section{Results}
We evaluated the performance of all REFINE agents offline, in a controlled user study, and in an unconstrained setting (RQ1). We also analyzed learners' interaction behaviour (RQ2) and how feedback shaped their conversations (RQ3).

\subsection{Performance of REFINE models (RQ1)}
We evaluated the quality of feedback reports and interactive conversations on the curated datasets and assessed their transferability to new real-world contexts.

\begin{figure}[t]
    \centering
    \includegraphics[width=\linewidth]{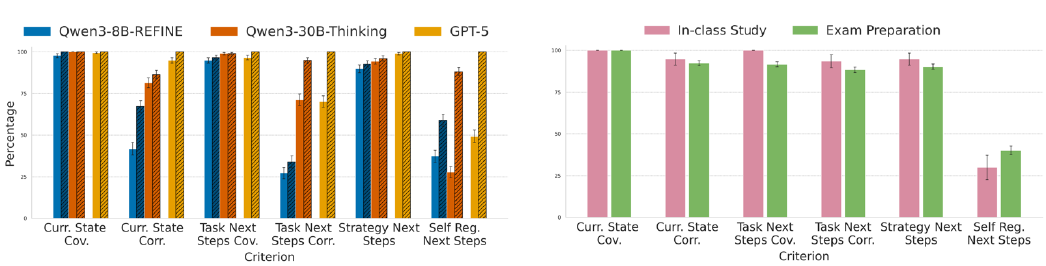}
    \caption{
    \textbf{(Left)} Percentage of positive Feedback Judge judgments on $D_{fb}$ across rubric dimensions before (solid bars) and after (hashed bars) a judge-guided feedback refinement step. \textbf{(Right)} Percentage of positive Feedback Judge judgments on $\mathcal{D}_{study}^{feedback}$ and $\mathcal{D}_{prep}^{feedback}$ with feedback generation model (Qwen3-30B-Thinking) after one iteration on Current-State and Task Next Steps correctness.}
    \label{fig:loop-criterion-bars}
    \vspace{-5mm}
\end{figure}

\vspace{1pt} \noindent \textbf{Feedback Generation}.
Figure~\ref{fig:loop-criterion-bars} (left) shows the effect of one judge-guided refinement step on feedback quality on $\mathcal{D}_{fb}$, where $t{=}0$ is the initial feedback and $t{=}1$ the first regeneration. Across models, self-regulated next steps improved most. Paired McNemar exact tests ($t{=}0$ vs.\ $t{=}1$, Benjamini–Hochberg corrected) confirmed significant gains for GPT-5, Qwen3-30B-Thinking, and Qwen3-8B-REFINE ($p<.001$). In addition, current-state correctness improved significantly for Qwen3-30B-Thinking ($p<.05$) and Qwen3-8B-REFINE ($p<.001$), while other criteria showed smaller or non-significant changes after one iteration. Clarity and praise received positive Feedback Judge judgments for all generated feedback across models and conditions from $t{=}0$. These results show that even a single targeted refinement step yields substantial improvements in correctness, supporting the use of a single iteration in deployment. 

Next, we analyzed how many refinement iterations were needed to achieve a perfect score on $\mathcal{D}_{fb}$. GPT-5 reached 100\% on all rubric criteria by $t{=}6$, with some criteria saturating earlier, and Qwen3-30B-Thinking converged by $t{=}5$. In contrast, Qwen3-8B-REFINE did not converge within $t \leq 20$, with task-level next-step correctness limiting performance (96.3\% at $t{=}20$).

Finally, we assessed whether feedback quality transfers beyond the offline setting. Figure~\ref{fig:loop-criterion-bars} (right) reports positive Feedback Judge judgments for the feedback generation model (Qwen3-30B-Thinking) on the in-class study dataset $\mathcal{D}_{study}^{feedback}$ and exam-preparation dataset $\mathcal{D}_{prep}^{feedback}$ after a single generation targeting current-state correctness and task-level next-step correctness. On $\mathcal{D}_{study}^{feedback}$, current-state and task-level next-step correctness were judged positive in over 93\% of cases, with full coverage and consistently positive judgments for clarity and praise; strategy-level next steps were often judged positive, while self-regulated next steps were less often positive. A similar pattern held on $\mathcal{D}_{prep}^{feedback}$: positive judgments remained high for current-state and task-level next-step criteria and strategy-level guidance, while self-regulated next steps were less often positive. This reflects our deployment set-up, where refinement was limited to current-state and task-level correctness to ensure real-time performance.

\begin{tcolorbox}[
  colback=gray!10, 
  colframe=gray!50, 
  boxrule=0.5pt, 
  arc=2pt,
  fontupper=\small
]
Feedback quality improves after Feedback LLM judge-guided refinement and remains robust when deployed in classroom and exam-preparation contexts.
\end{tcolorbox}

\begin{figure}[t]
    \centering
    \includegraphics[width=\linewidth]{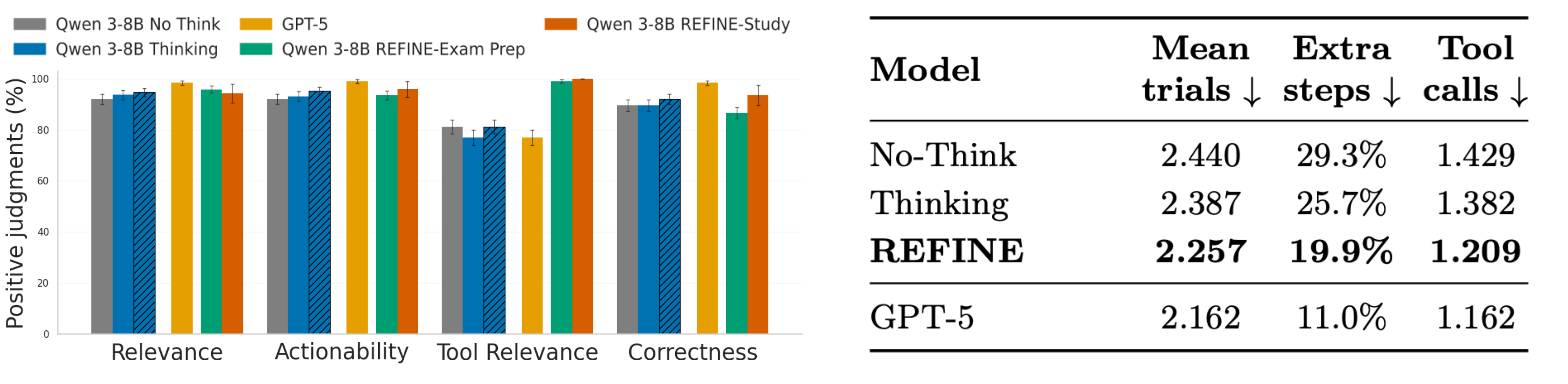}
    \caption{(\textbf{Left}) Percentage of positive judgments on dataset $\mathcal{D}_{int}^{test}$ by evaluation criterion. Qwen3-8B REFINE is hashed. (\textbf{Right}) Tool-mediated interaction efficiency on $\mathcal{D}_{int}^{test}$. Extra-step rate is the proportion of queries requiring more than the nominal two-step trajectory.} 
    \label{fig:judge_results}
    \vspace{-7mm}
\end{figure}

\vspace{1pt} \noindent \textbf{Interactive Feedback}.
Figure~\ref{fig:judge_results} (left) shows the Interactive Feedback Judge outcomes on the $\mathcal{D}_{int}^{test}$ dataset. Paired McNemar exact tests with Benjamini\allowbreak--\allowbreak Hochberg correction indicate that GPT-5 significantly outperforms Qwen~3-8B Thinking on actionability and correctness, with no significant differences in relevance or tool relevance. After fine-tuning on $\mathcal{D}_{int}^{train}$ under the REFINE interaction regime, this gap narrowed: Qwen~3-8B REFINE is not significantly different from GPT-5 on relevance, actionability, or tool relevance, although a significant gap in correctness remains. Overall, fine-tuning on tool-mediated interaction trajectories improves interactive feedback quality beyond reasoning-enabled base models and narrows the gap to GPT-5.

Figure~\ref{fig:judge_results} (left) also reports deployment outcomes from the Interactive Feedback Agent. On the in-class study dataset $\mathcal{D}_{study}^{int}$, responses received positive judgments of 96.0\% actionability, 94.4\% relevance, 93.6\%  correctness, and 100\% tool relevance. On the exam-preparation dataset $\mathcal{D}_{prep}^{int}$, performance remained similarly high (95.9\% relevance, 93.6\% actionability, 99.1\% tool relevance, and 86.7\% correctness), indicating that interactive feedback quality generalises to both classroom and unconstrained exam-preparation use.

Finally, we assessed tool efficiency of the different models on $\mathcal{D}_{int}^{test}$ (see Fig~\ref{fig:judge_results} (right)). Qwen3-8B REFINE reached final answers in fewer reasoning–tool steps than the thinking baseline (2.26 vs. 2.39 mean steps; paired Wilcoxon $p = 0.043$) and reduced the share of queries exceeding the nominal two-step trajectory (19.9\% vs. 25.7\%). GPT-5 sets a lower-friction ceiling (2.16 mean steps), with REFINE closing part of the gap. Overall, fine-tuning on REFINE interaction traces improves tool-use efficiency beyond enabling reasoning alone.

\begin{tcolorbox}[
  colback=gray!10, 
  colframe=gray!50, 
  boxrule=0.5pt, 
  arc=2pt,
  fontupper=\small
]
 REFINE yields more effective and efficient interactive feedback than reasoning-enabled base models, narrowing the gap to GPT-5 and transferring reliably to classroom and exam-preparation use.

\end{tcolorbox}
\vspace{-6mm}

\subsection{Student Engagement with Interactive Feedback (RQ2)}
We analysed student engagement by examining classroom follow-up question themes, relating them to task performance, and comparing them with unconstrained exam-preparation use.

\begin{figure}[t]
    \centering
    \includegraphics[width=\linewidth]{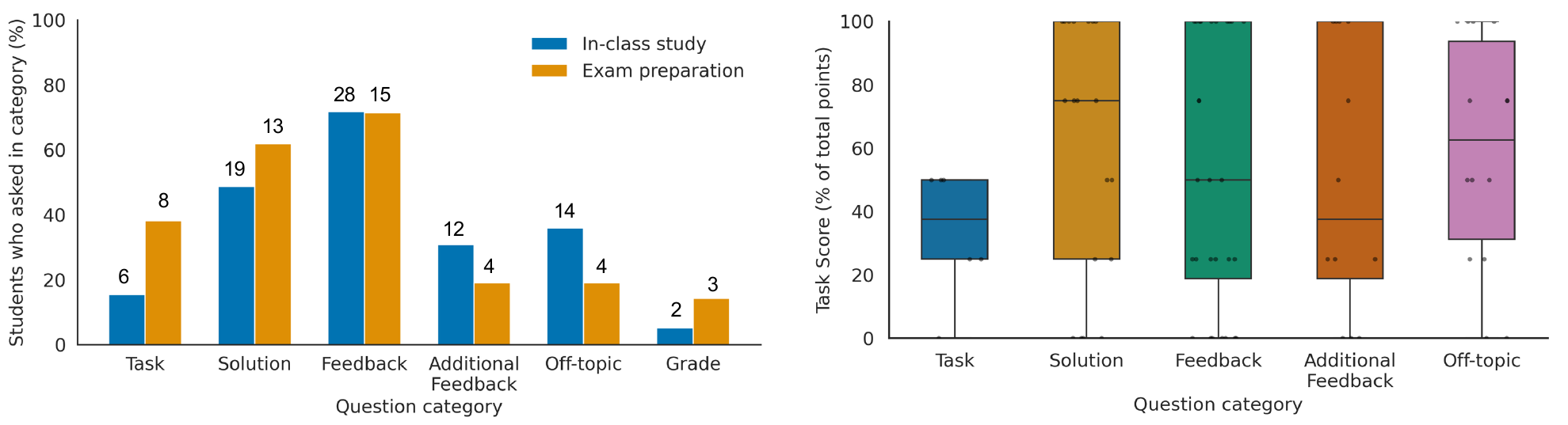}
  \caption{\textbf{(Left)} Question category use. Bars show the percentage of students who asked at least one question in category (in-class: $n{=}39$, exam-preparation: $n{=}21$). Counts above bars show number of students per condition. \textbf{(Right)} Distribution of task scores for students who asked at least one question in each category. }
  \label{fig:score-and-category-comparison}
    \vspace{-7mm}
\end{figure}

\vspace{1pt} \noindent \textbf{Classroom Study.}
We analysed student follow-up questions from the interaction dataset $\mathcal{D}_{study}^{int}$. Figure~\ref{fig:score-and-category-comparison} (left) shows the proportion of students who asked at least one question in each category. \textit{Feedback}-focused questions were most common, reflecting active engagement with the generated feedback and primarily involving requests for clarification, elaboration, or interpretation, and less frequently reassessment. \textit{Solution}-focused questions were also frequent and typically sought reassurance of correctness or a model solution for comparison.\textit{Task}-level questions were less common and mainly reflected difficulty understanding how to begin or what the problem required. \textit{Additional Feedback} questions extended beyond the immediate task, including requests to generalise learning or further strengthen acceptable solutions. \textit{Off-topic} questions largely concerned system usability or accessibility, including personal learning needs.

Next, we examined how question categories related to student performance (Figure~\ref{fig:score-and-category-comparison}, right). Students who asked \textit{Task} questions tended to have lower scores, consistent with difficulty interpreting problem requirements. In contrast, \textit{Solution} questions were associated with higher scores, reflecting engagement focused on verification rather than initial understanding. \textit{Feedback} and \textit{Additional Feedback} questions spanned a wide score range, indicating engagement by both lower- and higher-performing students seeking clarification or generalisation. \textit{Off-topic} questions occurred across performance levels and often concerned usability or accessibility needs, such as reading or writing difficulties. There were no significant differences in practice scores by question category using a student-level permutation test with overlapping category membership.

Finally, post-interaction perception survey responses consistently show positive ratings across all dimensions, with high mean scores: correctness ($4.1 \pm 0.8$), relevance ($4.2 \pm 0.7$), usefulness ($3.9 \pm 0.6$), and actionability ($3.9 \pm 0.8$).

\begin{tcolorbox}[
  colback=gray!10, 
  colframe=gray!50, 
  boxrule=0.5pt, 
  arc=2pt,
  fontupper=\small
]
Follow-up questions cluster into engagement categories: feedback- and solution-focused dominate, and task-level questions are more common at lower scores.
\end{tcolorbox}
\vspace{-1mm}

\vspace{1pt} \noindent \textbf{Unconstrained Exam-Preparation Use.}
We analysed student engagement with interactive feedback during unconstrained exam preparation using logs from $\mathcal{D}_{prep}^{feedback}$ and $\mathcal{D}_{prep}^{int}$. Students generated 362 feedback reports, with 99 (27.3\%) including at least one follow-up question, suggesting selective engagement only when additional clarification was needed.

Figure \ref{fig:score-and-category-comparison} (left) shows the percentage of students who asked at least one question in each category. Similar to the controlled classroom use, feedback-focused questions remained the most prevalent, with over 70\% of students engaging directly with the generated feedback, indicating sustained engagement beyond the classroom. Relative to the in-class study, exam-preparation use showed a higher prevalence of task- and solution-oriented questions, while off-topic and additional-feedback questions were more common in the classroom setting. A global permutation test did not detect a significant difference in category prevalence between the two settings. Qualitatively, unconstrained exam-preparation interactions exhibited a broader range of affective expressions, with students more frequently articulating frustration, reflecting on identified mistakes, and seeking confirmation or full solutions.

\vspace{-2mm}
\begin{tcolorbox}[
  colback=gray!10, 
  colframe=gray!50, 
  boxrule=0.5pt, 
  arc=2pt,
  fontupper=\small
]
During exam preparation, students engaged selectively with interactive feedback. Engagement themes closely mirrored those observed in classroom study.
\end{tcolorbox}

\vspace{-4mm}


\subsection{Impact of feedback content on follow-up questions (RQ3)}
\vspace{-1mm}
As a proxy for feedback uptake, we examined how feedback content in terms of highlighted specific rubric components influenced subsequent interactions. We fit a logistic regression predicting discussion of a rubric component in the chat as a function of whether that component was highlighted in the feedback, with standard errors clustered by student to account for repeated observations per student, $\mathrm{logit}(\Pr({discussed}))=\beta_0+\beta_1\, {highlighted}+\alpha_{{component}}$. Components highlighted in the feedback were substantially more likely to be discussed than non-highlighted components (OR = 8.10, 95\% CI [2.92, 22.46], $p < .001$), after accounting for differences across rubric components and repeated observations per student. Table~\ref{tab:feedback-steering} reports the corresponding proportions by rubric component, showing consistently higher discussion rates when components were highlighted.

\begin{table}[t]
\centering
\footnotesize
\setlength{\tabcolsep}{4pt}
\caption{Percentage of students who discussed a rubric component in chat when it was explicitly highlighted in the feedback (Discussed when highlighted) versus when it was not mentioned (Discussed when not highlighted).}
\begin{tabular}{lcc}
\toprule
\textbf{Rubric Component} & \textbf{Discussed (highlighted)} & \textbf{Discussed (not highlighted)} \\
\midrule
Basis & 89\% & 55\% \\
IH    & 86\% & 56\% \\
Step  & 92\% & 40\% \\
\midrule
\textbf{All} & 89\% & 51\% \\
\bottomrule
\end{tabular}
\label{tab:feedback-steering}
\vspace{-6mm}
\end{table}

\begin{tcolorbox}[
  colback=gray!10, 
  colframe=gray!50, 
  boxrule=0.5pt, 
  arc=2pt,
  fontupper=\small
]
Students were more likely to ask about  components highlighted in the feedback.
\end{tcolorbox}

\vspace{-3mm}
\section{Discussion and Conclusion}
\vspace{-1mm}
We introduced REFINE, a locally deployable multi-agent interactive feedback pipeline that combines an LLM-judge-guided feedback generation agent with a tool-calling interactive agent for follow-up conversations. Across offline benchmarks, a controlled classroom study, and unconstrained exam-preparation use, judge-guided refinement improves key pedagogical aspects while maintaining robust correctness under real-time constraints, and REFINE fine-tuning of a small thinking model narrows the gap to much larger GPT-5 on relevance and actionability while improving tool-use efficiency. Analyses of student interactions show that follow-up questions cluster into distinct engagement themes and that students are more likely to ask about rubric components highlighted in the feedback, suggesting that feedback can steer subsequent dialogue. While this does not directly measure learning gains, it provides evidence of how feedback influences student attention and interaction. This pattern also suggests that interactive feedback is not uniformly used across contexts, but is instead invoked when learners encounter uncertainty, highlighting the need for systems that support both lightweight one-shot feedback and optional deeper interaction.

\textbf{Limitations.} Our deployments were conducted in a single course and institution, and we focus on interactional evidence rather than direct learning gains. To meet latency constraints in deployment, feedback refinement targeted only a subset of rubric criteria, which likely underestimates performance on dimensions not iterated at inference time. In the classroom study, students were required to ask at least three follow-up questions, which may have increased interaction volume relative to the unconstrained exam-preparation setting.

\textbf{Future work.} Next steps focus on connecting interactive traces and feedback uptake to downstream learning outcomes. Additionally, future systems should explore adaptive, instructor-steerable refinement policies that decide which rubric dimensions to iterate based on student needs and instructional intent, and extend deployment to additional courses and domains with richer tools and content.

\begin{credits}
\subsubsection{\ackname}
This project was substantially financed by the Swiss State Secretariat for Education, Research and Innovation (SERI). We thank Rüdiger Urbanke for his support in conducting the study and data collection, and Bahar Radmehr and Bianca Pitu for their support during the user study.

\subsubsection{\discintname}
The authors have no competing interests to declare that are relevant to the content of this article.
\end{credits}

\FloatBarrier

\bibliographystyle{splncs04}
\bibliography{bib}

\end{document}